\documentclass[pre,twocolumn,twoside,showpacs,byrevtex,superscriptaddress]{revtex4}

\usepackage{currfile}

\usepackage{graphicx,epsfig,verbatim,enumerate}
\usepackage{amssymb,amsmath,mathdots}
\usepackage{ifthen,centernot}
\usepackage{multirow,lipsum,hyperref}

\hypersetup{
        colorlinks = true,
        allcolors = blue
}

\usepackage{tikz-cd}
\usetikzlibrary{arrows}
\tikzset{
  commutative diagrams/.cd,
  arrow style=tikz,
  diagrams={>=space}}

\newboolean{twocolswitch}

\newcommand{\sindex}[1]{}
\newcommand{\nindex}[1]{}

\newcommand{\www}[1]{\url{#1}}

\usepackage{lettrine}

\usepackage{color}

\RequirePackage{color}\definecolor{RED}{rgb}{1,0,0}\definecolor{BLUE}{rgb}{0,0,1} 

\setboolean{twocolswitch}{true}

\begin{document}

\title{

Identifying missing dictionary entries with frequency-conserving context models
}

\author{
\firstname{Jake Ryland}
\surname{Williams}
}

\email{jake.williams@uvm.edu}

\affiliation{Department of Mathematics \& Statistics,
  Vermont Complex Systems Center,
  Computational Story Lab,
  \& the Vermont Advanced Computing Core,
  The University of Vermont,
  Burlington, VT 05401.}

\author{
\firstname{Eric M.}
\surname{Clark}
}

\email{eric.clark@uvm.edu}

\affiliation{Department of Mathematics \& Statistics,
  Vermont Complex Systems Center,
  Computational Story Lab,
  \& the Vermont Advanced Computing Core,
  The University of Vermont,
  Burlington, VT 05401.}

\author{
\firstname{James P.}
\surname{Bagrow}
}
\email{james.bagrow@uvm.edu}

\affiliation{Department of Mathematics \& Statistics,
  Vermont Complex Systems Center,
  Computational Story Lab,
  \& the Vermont Advanced Computing Core,
  The University of Vermont,
  Burlington, VT 05401.}

\author{
\firstname{Christopher M.}
\surname{Danforth}
}
\email{chris.danforth@uvm.edu}

\affiliation{Department of Mathematics \& Statistics,
  Vermont Complex Systems Center,
  Computational Story Lab,
  \& the Vermont Advanced Computing Core,
  The University of Vermont,
  Burlington, VT 05401.}

\author{
\firstname{Peter Sheridan}
\surname{Dodds}
}
\email{peter.dodds@uvm.edu}

\affiliation{Department of Mathematics \& Statistics,
  Vermont Complex Systems Center,
  Computational Story Lab,
  \& the Vermont Advanced Computing Core,
  The University of Vermont,
  Burlington, VT 05401.}

\date{\today}

\begin{abstract}
  In an effort to better understand 
meaning from natural language texts,
we explore methods aimed at 
organizing lexical objects into contexts.
A number of these methods for organization
fall into a family defined by word ordering.
Unlike demographic or spatial partitions of data,
these collocation models are of special importance
for their universal applicability.
While we are interested here in text
and have framed our treatment appropriately,
our work is potentially applicable to other areas of research
(e.g., speech, genomics, and mobility patterns)
where one has ordered categorical data,
(e.g., sounds, genes, and locations).
Our approach focuses on the phrase 
(whether word or larger)
as the primary meaning-bearing lexical unit
and object of study.
To do so, we employ our previously developed framework
for generating word-conserving phrase-frequency data.
Upon training our model with the Wiktionary---an extensive, online, collaborative, and open-source dictionary
that contains over $100,000$ phrasal-definitions---we develop highly effective filters for the
identification of meaningful, missing phrase-entries.
With our predictions
we then engage the editorial 
community of the Wiktionary
and propose short lists 
of potential missing entries for definition,
developing a breakthrough, lexical extraction technique,
and expanding our knowledge of
the defined English lexicon of phrases.

\end{abstract}


\maketitle

\section{Background}
Starting with the work of Shannon \cite{shannon1948a},
information theory has grown enormously
and has been shown by Jaynes
to have deep connections to statistical mechanics~\cite{jaynes1957a}.
We focus on a particular aspect of Shannon's work,
namely joint probability distributions between word-types
(denoted $w\in W$),
and their groupings by appearance-orderings, 
or, \emph{contexts}
(denoted $c\in C$).
For a word appearing in text, 
Shannon's model assigned context
according to the word's immediate antecedent.
In other words, the sequence 
$$\cdots\:w_{i-1}\:w_{i}\:\cdots$$
places this occurrence of the word-type of $w_{i}$ 
in the context of $w_{i-1}\:\star$
(uniquely defined by the word-type of $w_{i-1}$),
where ``$\star$'' denotes  ``any word''.
This experiment was novel, 
and when these transition probabilities were observed,
he found a method for the automated production of language that
far better resembled true English text
than simple adherence to relative word frequencies.

Later, though still early on
in the history of modern
computational linguistics and
natural language processing,
theory caught up with Shannon's work.
In 1975, Becker wrote~\cite{becker1975a}:
\begin{quote}
  My guess is that phrase-adaption and
  generative gap-filling are very roughly
  equally important in language production, as
  measured in processing time spent on each,
  or in constituents arising from each. One
  way of making such an intuitive estimate is
  simply to listen to what people actually say
  when they speak. An independent way of
  gauging the importance of the phrasal
  lexicon is to determine its size. 
\end{quote}
Since then, with the rise of computation
and increasing availability of electronic text,
there have been numerous extensions of Shannon's context model.
These models have generally been information-theoretic applications as well,
mainly used to predict word associations~\cite{church1990a} 
and to extract multi-word expressions (MWEs)~\cite{smadja1993a}.
This latter topic has been one of extreme importance
for the computational linguistics community~\cite{ramisch2014a},
and has seen many approaches aside from the information-theoretic,
including with part-of-speech taggers~\cite{justeson1995a}
(where categories, e.g., noun, verb, etc. are used to identify word combinations)
and with syntactic parsers~\cite{seretan2008a}
(where rules of grammar are used to identify word combinations).
However, almost all of these methods 
have the common issue of scalability~\cite{pecina2010a},
making them difficult to use for the extraction of phrases
of more than two words.

Information-theoretic extensions of 
Shannon's context model 
have also been used 
by Piantadosi et al.~\cite{piantadosi2011a}
to extend the work of Zipf~\cite{zipf1935a},
using an entropic derivation called 
the Information Content (IC):
\begin{equation}
  I(w)=-\sum_{c\in C}P(c\mid w)\log P(w\mid c)
  \label{eq:IC}
\end{equation}
and measuring its associations to word lengths.
Though there have been concerns
over some of the conclusions reached in this work~\cite{reilly2011a,piantadosi2011b,cancho2012a,piantadosi2013a},
Shannon's model was somewhat generalized, 
and applied to $3$-gram, $4$-gram and $5$-gram
context models to predict word lengths.
This model was also used by Garcia et al.~\cite{garcia2012a} 
to assess the relationship between
sentiment (surveyed emotional response) norms and IC
measurements of words.
However their application of the formula
\begin{equation}
  I(w)=-\frac{1}{f(w)}\sum_{i=1}^{f(w)}\log P(w\mid c_i),
\end{equation}
to $N$-grams data
was wholly incorrect,
as this special representation applies only
to corpus-level data, i.e., 
plot line-human readable text,
and \emph{not} the frequency-based $N$-grams.

In addition to the above considerations,
there is also the general concern of non-physicality
with imperfect word frequency conservation, 
which is exacerbated by the Piantadosi et al.
extension of Shannon's model.
To be precise, 
for a joint distribution of words
and contexts that is \emph{physically} 
related to the appearance of words 
on ``the page'',
there should be conservation
in the marginal frequencies:
\begin{equation}
  f(w)=\sum_{c\in C}f(w,c),
\end{equation}
much like that discussed in~\cite{church1990a}.
This property is not upheld using any 
true, sliding-window $N$-gram data 
(e.g.,~\cite{google2006a,michel2011a,lin2012a}).
To see this, we recall that in 
both of~\cite{garcia2012a} and \cite{piantadosi2011a}, 
a word's $N$-gram context
was defined by its immediate $N-1$ antecedents.
However, by this formulation we note that the first word of
a page
appears as \emph{last} in no $2$-gram, 
the second appears as \emph{last} in no $3$-gram, 
and so on.

These word frequency misrepresentations 
may seem to be of little importance at the text or page level,
but since the methods for large-scale $N$-gram parsing
have adopted the practice of stopping at
sentence and clause boundaries~\cite{lin2012a},
word frequency misrepresentations (like those discussed above)
have become very significant.
In the new format, 
$40\%$ of the words in a 
sentence or clause
of length five
have no $3$-gram context (the first two).
As such, when these context models 
are applied to modern $N$-gram data,
they are incapable of accurately representing
the frequencies of words expressed.
We also note that despite the advances in
processing made in the construction of the
current Google $N$-grams corpus~\cite{lin2012a}, 
other issues have been found,
namely regarding the source texts taken~\cite{pechenick2015a}.

The $N$-gram expansion of Shannon's model incorporated
more information on relative word placement,
but perhaps an ideal scenario would arise when
the frequencies of author-intended phrases are exactly known.
Here, one can conserve word frequencies
(as we discuss in section II)
when a context for an instance of a word
is defined by its removal pattern,
i.e., the word
``cat''
appears in the context
``$*$ in the hat'',
when the phrase
``cat in the hat''
is observed.
In this way, a word-type appears in
as many contexts as there are phrase-types
that contain the word.
While we consider the different phrase-types
as having rigid and different meanings,
the words underneath can be looked at as
having more flexibility, often in need of disambiguation.
This flexibility is quite similar to an aspect of
a physical model of lexicon learning~\cite{reisenauer2013a},
where a ``context size'' control parameter
was used to tune the number of plausible but unintended meanings
that accompany a single word's true meaning.
An enhanced model of lexicon learning that focuses
on meanings of phrases could then explain the need for
disambiguation when reading by words.

We also note that 
there exist many other methods for grouping occurrences
of lexical units to produce informative context models. 
As early as $1992$~\cite{resnik1992a}, 
Resnik showed class categorizations of words
(e.g., verbs and nouns)
could be used to produce informative
joint probability distributions.
In 2010, Montemurro et al.~\cite{montemurro2010a}
used joint distributions of words and
arbitrary equal-length parts of texts 
to entropically quantify the
semantic information encoded in 
written language.
Texts tagged with metadata like
genera~\cite{dodds2009b}, 
time~\cite{dodds2011f}, 
location~\cite{mitchell2013a}, and
language~\cite{dodds2015a},
have rendered straightforward and clear examples
of the power in a (word-frequency conserving) joint pmf, 
at shedding light on social phenomena
by relating words to classes.
Additionally, while their work did not leverage word frequencies
or the joint pmf's possible, Benedetto et al.~\cite{benedetto2002a}
used metadata of texts to train
language and authorship detection algorithms, and further,
construct accurate phylogenetic-like trees
through application of compression distances. 
Though metadata approaches to context are informative,
with their power there is simultaneously
a loss of applicability
(metadata is frequently not present), 
as well as a loss of bio-communicative relevance
(humans are capable of inferring social information from text in isolation).

\section{Frequency-conserving context models}

\begin{table}[t!]  
   \begin{center}
     \scalebox{0.89}{
     \begin{tabular}{|c|c|c|c|c|c|}
       \hline
       phrase & $\ell(s_{i\cdots j})=1$ & $\ell(s_{i\cdots j})=2$ & $\ell(s_{i\cdots j})=3$ & $\ell(s_{i\cdots j})=4$ & $\cdots$ \\\hline
       $w_1$ & $\star$ & - & - & - & $\cdots$ \\\hline 
       $w_1\:w_2$ & $\star\:w_2$ & $\star\:\star$ & - & - & $\cdots$ \\ 
       & $w_1\:\star$ &                &   &   & $\cdots$ \\\hline
       $w_1\:w_2\:w_3$ & $\star\:w_2\:w_3$ & $\star\:\star\:w_3$ & $\star\:\star\:\star$ & - & $\cdots$ \\ 
       & $w_1\:\star\:w_3$ & $w_1\:\star\:\star$ &                       &   & $\cdots$ \\
       & $w_1\:w_2\:\star$ &                   &                       &   & $\cdots$ \\\hline
       
       $w_1\:w_2\:w_3\:w_4$ & $\star\:w_2\:w_3\:w_4$ & $\star\:\star\:w_3\:w_4$ & $\star\:\star\:\star\:w_4$ & $\star\:\star\:\star\:\star$ & $\:\cdots\:$ \\ 
       & $w_1\:\star\:w_3\:w_4$ & $w_1\:\star\:\star\:w_4$ & $w_1\:\star\:\star\:\star$ &                              & $\cdots$ \\
       & $w_1\:w_2\:\star\:w_4$ & $w_1\:w_2\:\star\:\star$ &                          &                              & $\cdots$ \\
       & $w_1\:w_2\:w_3\:\star$ &                      &                          &                              & $\cdots$ \\\hline
       
       $\vdots$ & $\vdots$ & $\vdots$ & $\vdots$ & $\vdots$& $\ddots$ \\\hline
       
     \end{tabular}
     }
     \caption{
       A table showing the expansion of context lists 
       for longer and longer phrases.
       We define the internal contexts of phrases 
       by the removal of individual sub-phrases.
       These contexts are represented as phrases 
       with words replaced by $\star$'s.
       Any phrases whose word-types match
       after analogous sub-phrase removals 
       share the matching context.
       Here, the columns are labeled $1$--$4$ by
       sub-phrase length. 
     }
     \label{tab:conTab}
   \end{center}
   
\end{table}

In previous work~\cite{williams2014a} we developed 
a scalable and general framework for 
generating frequency data for $N$-grams,
called random text partitioning. 
Since a phrase-frequency distribution, $S$, 
is balanced with regard to its 
underlying word-frequency distribution, $W$,
\begin{equation}
  \sum_{w\in W}f(w) = \sum_{s\in S} \ell(s)f(s)
\end{equation}
(where $\ell$ denotes the phrase-length norm,
 which returns the length of a phrase in numbers of words)
it is easy to produce a 
symmetric generalization 
of Shannon's model that integrates 
all phrase/$N$-gram lengths and 
all word placement/removal points.
To do so, we define $W$ and $S$ to be the 
sets of words and 
(text-partitioned) 
phrases from a text respectively,
and let $C$ be the collection of all single word-removal
patterns from the phrases of $S$.
A joint frequency, $f(w,c)$,
is then defined by
the partition frequency
of the phrase that is formed 
when $c$ and $w$ are composed.
In particular, 
if $w\text{ composed with }c\text{ renders }s$, 
we then set $f(w,c) = f(s)$,
which produces a context model on the words
whose marginal frequencies
preserve their original frequencies from 
``the page.''
In particular we refer to this,
or such a model for phrases, 
as an `external context model,'
since the relations are produced by structure
external to the semantic unit.

It is good to see the external word-context
generalization emerge, 
but our interest actually lies in the development
of a context model for the phrases themselves.
To do so, 
we define the `internal contexts'
of a phrase
by the patterns generated 
through the removal of sub-phrases.
To be precise, 
for a phrase $s$,
and a sub-phrase $s_{i\cdots j}$
ranging over words $i$ through $j$,
we define the context 
\begin{equation}
  c_{i\cdots j} = w_1\:\cdots\:w_{i-1}\:\star\:\cdots\:\star\:w_{j+1}\:\cdots\:w_{\ell(s)}
\end{equation}
to be the collection of same-length phrases
whose analogous word removal ($i$ through $j$)
renders the same pattern
(when word-types are considered).
We present the contexts
of generalized phrases of lengths $1$--$4$ 
in Tab.~\ref{tab:conTab}, as described above.
Looking at the table,
it becomes clear that
these contexts are actually
a mathematical formalization
of the generative gap filling
proposed in \cite{becker1975a},
which was semi-formalized
by the phrasal templates
discussed at length by Smadja et al. in~\cite{smadja1993a}.
Between our formulation and that of Smadja, 
the main difference of definition lies in
our restriction to contiguous word sequence 
(i.e., sub-phrase) removals,
as is necessitated by 
the mechanics of the secondary partition process,
which defines the context lists.

The weighting of the contexts
for a phrase is accomplished
simultaneously with their definition
through a secondary partition process
describing the inner-contextual modes 
of interpretation for the phrase.
The process is as follows.
In an effort to relate an observed
phrase to other known phrases,
the observer selectively ignores
a sub-phrase of the original phrase.
To retain generality,
we do this by considering 
the \emph{random} partitions 
of the original phrase,
and then assume that a sub-phrase is
ignored from a partition 
with probability proportional 
to its length,
to preserve word 
(and hence phrase) frequencies.
The conditional probabilities of 
inner context are then:
\begin{equation}
  \begin{split}
    P&(c_{i\cdots j}\mid s)=\\
    P&(\text{ignore }s_{i\cdots j}\text{ given a partition of }s)=\\
    P&(\text{ignore }s_{i\cdots j}\text{ given }s_{i\cdots j}\text{ is partitioned from }s)\times\\
    P&(s_{i\cdots j}\text{ is partitioned from }s).
  \end{split}
\end{equation}
Utilizing the partition probability and our assumption,
we note from our work in \cite{williams2014a} that
\begin{equation}
  \ell(s) = \sum_{1\leq i<j\leq \ell(s)}\ell(s_{i\cdots j}) P_q(s_{i\cdots j} \mid s),
\end{equation}
which ensures through defining
\begin{equation} 
  P(c_{i\cdots j}\mid s) = \frac{\ell(s_{i\cdots j})}{\ell(s)}P_q(s_{i\cdots j} \mid s),
\end{equation}
the production of a valid,
phrase-frequency preserving context model:
\begin{equation}
  \begin{split}
    \sum_{c\in C}f(c,s)=&\sum_{i<j\leq\ell(s)}P(c_{i\cdots j}\mid s)f(s)\\
    =&f(s)\sum_{1\leq i<j\leq\ell(s)}\frac{\ell(s_{i\cdots j})}{\ell(s)}P_q(s_{i\cdots j} \mid s) = f(s),
  \end{split}
\end{equation}
which preserves the underlying 
frequency distribution of phrases.
Note here that beyond this point in the document
we will used the normalized form,
\begin{equation}
  P(c,s)=\frac{f(c,s)}{\underset{s\in S}{\sum}\underset{c\in C}{\sum}f(c,s)},
\end{equation}
for convenience in the 
derivation of expectations 
in the next section.

\section{Likelihood of Dictionary Definition}
\label{sec:deflik}
In this section we exhibit the power
of the internal context model through 
a lexicographic application, 
deriving a measure of meaning and definition for phrases
with empirical phrase-definition data taken from 
a collaborative open-access dictionary~\cite{wiktionary2014}
(see Sec.~\ref{sec:matmet} for more information on our data and the Wiktionary).
With the rankings that this measure derives,
we will go on to propose phrases for definition with the 
editorial community of the Wiktionary in an ongoing live experiment,
discussed in Sec.~\ref{sec:misent}.

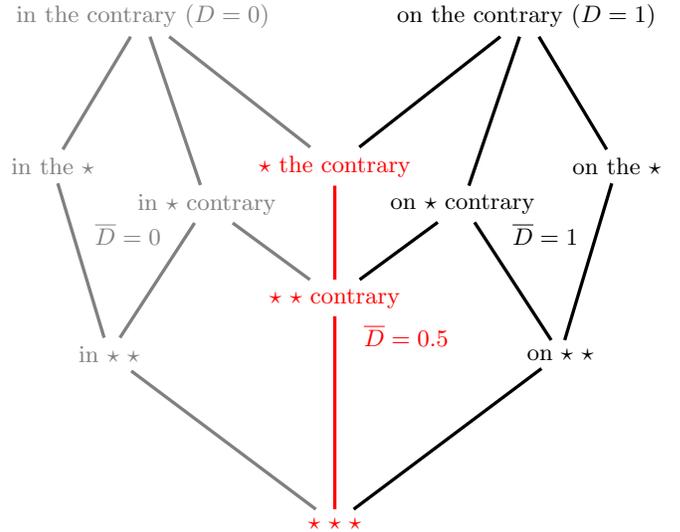
\begin{figure}[t!]
  \centering
  \begin{tikzpicture}[%
      back line/.style={densely dotted},
      cross line/.style={preaction={draw=white, -,line width=6pt}}]
    
    \node at (0,-0.25) (EPS) {\textcolor{red}{$\star$ $\star$ $\star$}};
    
    \node at (0,2) (LC) {};
    \node at (0,2.75) (LC2) {};
    
    \node [right of=LC2, node distance=0cm, red] (Wo) {$\star$ $\star$ contrary};
    \node [left of=LC, node distance=3cm, gray] (Ma) {in $\star$ $\star$};
    \node [right of=LC, node distance=3cm, black] (Fu) {on $\star$ $\star$};
    
    \node at (0,4) (MC) {};
    \node at (2.8,3.6) (pDefB1) {$\overline{D}=1$};
    \node [gray] at (-2.75,3.6) (pDefB1) {$\overline{D}=0$};
    \node [red] at (0.95,2.25) (pDefB1) {$\overline{D}=0.5$};
    \node at (0,4.5) (MC2) {};
    
    \node [above of=MC, node distance=0.5cm, red] (WiWo) {$\star$ the contrary};
    \node [left of=MC, node distance=1.7cm, gray] (MaWo) {in $\star$ contrary};
    \node [left of=MC2, node distance=3.75cm, gray] (MaWi) {in the $\star$};          
    \node [right of=MC2, node distance=3.75cm, black] (FuWi) {on the $\star$};
    \node [right of=MC, node distance=1.7cm, black] (FuWo) {on $\star$ contrary};
    
    \node at (0,6.5) (C) {};
    
    \node [left of=C, node distance=2.55cm, gray] (MaWiWo) {in the contrary ($D=0$)};
    \node [right of=C, node distance=2.55cm, black] (FuWiWo) {on the contrary ($D=1$)};
    
    \draw[cross line,-, gray, line width=1.25] (Wo) -- (MaWo);
    \draw[cross line,-, red, line width=1.25] (Wo) -- (WiWo);
    \draw[cross line,-, black, line width=1.25] (Wo) -- (FuWo);
    
    \draw[cross line,-, black, line width=1.25] (Fu) -- (FuWi);
    \draw[cross line,-, black, line width=1.25] (Fu) -- (FuWo);
    
    \draw[cross line,-, gray, line width=1.25] (Ma) -- (MaWo);
    \draw[cross line,-, gray, line width=1.25] (Ma) -- (MaWi);
    
    \draw[cross line,-, gray, line width=1.25] (WiWo) -- (MaWiWo);
    \draw[cross line,-, black, line width=1.25] (WiWo) -- (FuWiWo);
    
    \draw[cross line,-, gray, line width=1.25] (MaWi) -- (MaWiWo);
    \draw[cross line,-, gray, line width=1.25] (MaWo) -- (MaWiWo);
    \draw[cross line,-, black, line width=1.25] (FuWi) -- (FuWiWo);
    \draw[cross line,-, black, line width=1.25] (FuWo) -- (FuWiWo);
    
    \draw[cross line,-, gray, line width=1.25] (EPS) -- (Ma);
    \draw[cross line,-, black, line width=1.25] (EPS) -- (Fu);
    \draw[cross line,-, red, line width=1.25] (EPS) -- (Wo);
    
  \end{tikzpicture}  
  \caption{
    An example showing the sharing of contexts by similar phrases.
    Suppose our text consists of the two phrases,
    ``in the contrary'' and ``on the contrary'',
    and that each occurs once, 
    and that the latter has definition ($D=1$) while the former does not.
    In this event, we see that the three shared contexts:
    ``$\star~\star~\star$'', ``$\star~\star~\text{contrary}$'', and ``$\star~\text{the}~\text{contrary}$'',
    present elevated likelihood ($\overline{D}$) values,
    indicating that the phrase ``in~the~contrary'' may have
    meaning and be worthy of definition.
  }
  \label{fig:contextTree}
\end{figure}

To begin, we define the dictionary indicator, $D$, 
to be a binary norm on phrases,
taking value $1$ when a phrase appears in the dictionary,
(i.e., has definition)
and taking value $0$ when a phrase is unreferenced.
The dictionary indicator tells us when a phrase
has reference in the dictionary,
and in principle
can be replaced with other indicator norms,
for other purposes.
Moving forward, 
we note of an intuitive description of the distribution average:
\begin{equation*}
  \begin{split}
    \overline{D}(S)&=\sum_{t\in S}D(t)P(t) \\
    &= P(\text{randomly drawing a defined phrase from S}),
  \end{split}
\end{equation*}
and go on to derive an alternative expansion
through application of the context model:
\begin{equation}
  \begin{split}
  \overline{D}(S)
  &=\sum_{t\in S}D(t)P(t)\\  
        &=\sum_{t\in S}D(t)P(t)\sum_{c\in C}P(c\mid t)\sum_{s \in S}P(s\mid c)\\
  &=\sum_{c\in C}P(c)\sum_{t\in S}D(t)P(t\mid c)\sum_{s \in S}P(s\mid c)\\
  &=\sum_{c\in C}P(c)\sum_{s \in S}P(s\mid c)\sum_{t\in S}D(t)P(t\mid c)\\
  &=\sum_{s\in S}P(s)\sum_{c \in C}P(c\mid s)\sum_{t\in S}D(t)P(t\mid c)\\
  &=\sum_{s\in S}P(s)\sum_{c \in C}P(c\mid s)\overline{D}(c\mid S).
  \end{split}
\end{equation}
In the last line 
we then interpret:
\begin{equation}
 \overline{D}(C\mid s) = \sum_{c \in C}P(c\mid s)\overline{D}(c\mid S),
\end{equation}
to be the likelihood
(analogous to the IC equation presented here as equation~\ref{eq:IC})
that a phrase, which is 
randomly drawn from 
a context of $s$,
to have definition in the dictionary.
To be precise, 
we say $\overline{D}(C\mid s)$ 
is the likelihood of dictionary definition of 
the context model $C$,
given the phrase $s$,
or,
when only one $c\in C$ 
is considered, 
we say $\overline{D}(c\mid S) = \sum_{t\in S}D(t)P(t\mid c)$
is the likelihood of dictionary definition of 
the context $c$, given $S$.
Numerically, 
we note that the distribution-level values,
$\overline{D}(C\mid s),$
``extend'' the dictionary over all $S$,
smoothing out the binary data
to the full lexicon
(uniquely for phrases of more than one word,
which have no interesting space-defined
internal structure)
through the relations of the model.
In other words, 
though $\overline{D}(C\mid s)\neq 0$
may now only indicate the \emph{possibility} 
of a phrase having definition,
it is still a strong indicator,
and most importantly,
may be applied to never-before-seen expressions.
We illustrate the extension of the dictionary
through $\overline{D}$ in Fig.~\ref{fig:contextTree},
where it becomes clear that the topological structure
of the associated network of contexts is crystalline,
unlike the small-world phenomenon observed for the
words of a thesaurus in~\cite{motter2002a}.
However, this is not surprising,
given that the latter is a conceptual network
defined by common meanings,
as opposed to a rigid, physical property,
such as word order.

\section{Predicting missing dictionary entries}
\label{sec:misent}
Starting with the work of Sinclair~\cite{sinclair1987a}
(though the idea was proposed more than $10$ years earlier
by Becker in \cite{becker1975a}),
lexicographers have been building dictionaries 
based on language as it is spoken and written,
including idiomatic, slang-filled, and grammatical expressions
\cite{cobuild,wiktionary,urban,slang}.
These dictionaries have proven highly-effective
for non-primary language learners,
who may not be privy to cultural metaphors.
In this spirit, 
we utilize the context model derived above
to discover phrases that are undefined,
but which may be in need of definition
for their similarity to other, defined phrases.
We do this in a corpus-based way, 
using the definition likelihood $\overline{D}(C\mid s)$ 
as a secondary filter to frequency.
The process is in general quite straightforward,
and first requires a ranking of phrases
by frequency of occurrence, $f(s)$.
Upon taking the first $s_1,...,s_N$ 
frequency-ranked phrases
($N=100,000$, for our experiments),
we reorder the list 
according to the values $\overline{D}(C\mid s)$ (descending).
The top of such a double-sorted list
then includes phrases that are both frequent 
and similar to defined phrases.

With our double-sorted lists
we then record those phrases having
no definition or dictionary reference,
but which are at the top. 
These phrases are quite often meaningful
(as we have found experimentally, see below)
despite their lack of definition,
and as such we propose this method
for the automated generation of 
short lists for editorial investigation of definition.

\section{Materials and methods}
\label{sec:matmet}
For its breadth,
open-source nature,
and large editorial community,
we utilize dictionary data from
the Wiktionary~\cite{wiktionary2014}
(a Wiki-based open content dictionary)
to build the dictionary-indicator norm, 
setting $D(s)=1$
if a phrase $s$ has reference or redirect.

We apply our filter for
missing entry detection
to several large corpora from
a wide scope of content.
These corpora are:
twenty years of New York Times articles (NYT, 1987--2007)~\cite{times2008},
approximately $4\%$ of a year's tweets (Twitter, 2009)~\cite{twitter2009},
music lyrics from thousands of songs and authors (Lyrics, 1960--2007)~\cite{dodds2009b},
complete Wikipedia articles (Wikipedia, 2010)~\cite{wikipedia2010},
and Project Gutenberg eBooks collection (eBooks, 2012)~\cite{gutenberg2012} 
of more than $30,000$ public-domain texts.
We note that these are all unsorted texts,
and that Twitter, eBooks, Lyrics, and to an extent, Wikipedia 
are mixtures of many languages (though majority English).
We only attempt missing entry prediction
for phrase lengths ($2$--$5$),
for their inclusion in other major collocation corpora~\cite{lin2012a},
as well as their having the most data in the dictionary.
We also note that all text processed is taken lower-case.

\begin{figure}[t!]
    \includegraphics[width=0.495\textwidth]{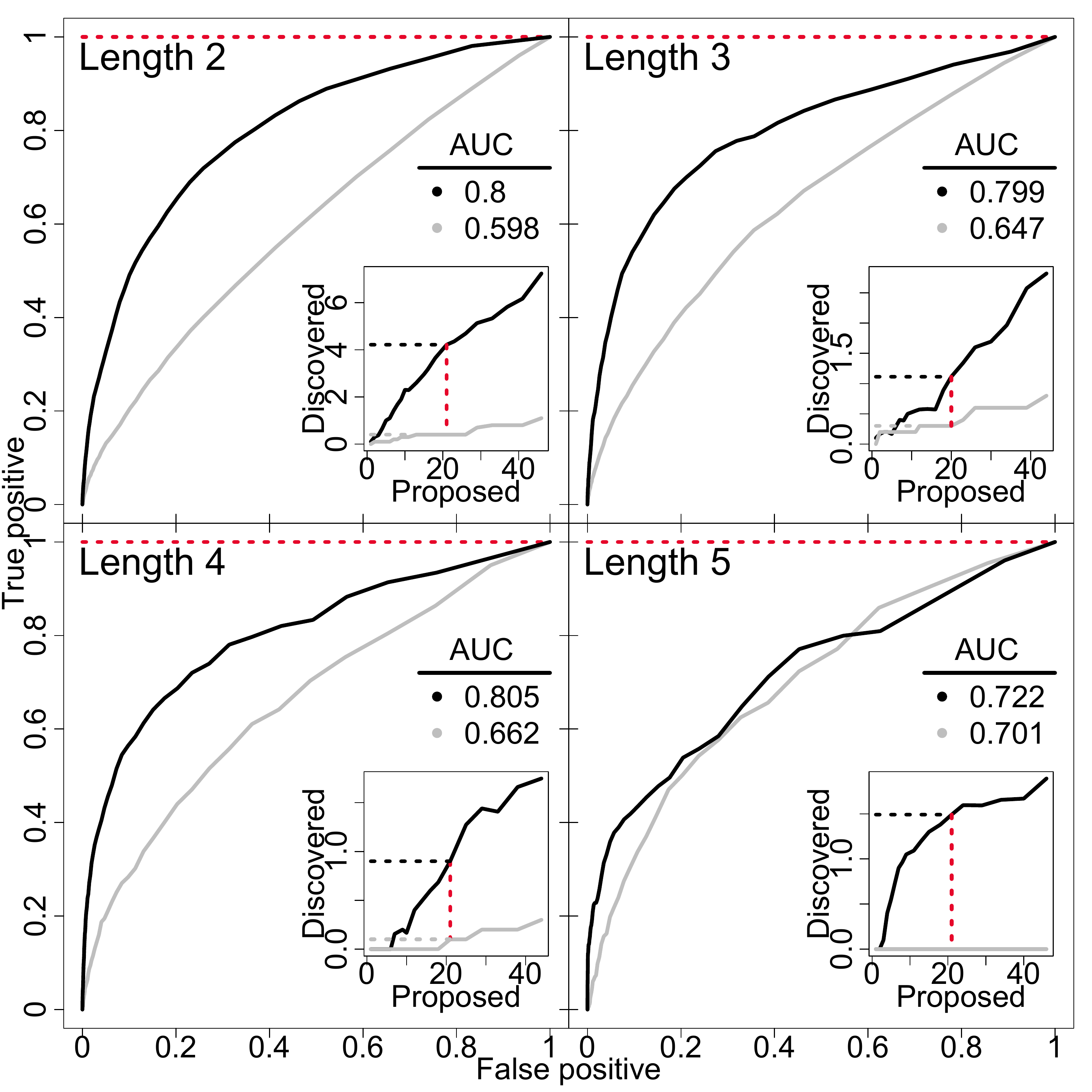}
  \caption{
    With data taken from the Twitter corpus, 
    we present ($10$-fold) cross-validation results
    for the filtration procedures.
    For each of the lengths $2$, $3$, $4$, and $5$,
    we show the ROC curves (\textbf{Main Axes}),
    comparing true and false positive rates
    for both the likelihood filters (black),
    and for the frequency filters (gray).
    There, we see increased performance 
    in the likelihood classifiers
    (except possibly for length $5$),
    which is reflected in the AUCs
    (where an AUC of $1$ indicates a perfect classifier).
    We also monitor the average number of 
    missing entries discovered
    as a function of the number of entries proposed (\textbf{Insets}),
    for each length.
    There, the horizontal dotted lines indicate 
    the average numbers
    of missing entries discovered for both the
    likelihood filters (black)
    and for the frequency filters (gray)
    when short lists of $20$ phrases were taken 
    (red dotted vertical lines).
    From this we see an indication
    that even the $5$-gram likelihood filter is
    effective at detecting missing entries in short lists,
    while the frequency filter is not.
  }
  \label{fig:twitter.crossval}
\end{figure}

To understand our results, 
we perform a 10-fold cross-validation on
the frequency and likelihood filters.
This is executed by random splitting the
Wiktionary's list of defined phrases
into $10$ equal-length pieces,
and then performing $10$ parallel experiments
In each of these experiments we determine
the likelihood values, $\overline{D}(C\mid s)$,
by a distinct $\frac{9}{10}$'s of the data.
We then order the union set of 
the $\frac{1}{10}$-withheld
and the Wiktionary-undefined
phrases by their likelihood 
(and frequency) values descending,
and accept some top segment of the list,
or, `short~list',
coding them as positive by the experiment.
For such a short list,
we then record
the true positive rates, i.e.,
portion of all $\frac{1}{10}$-withheld 
truly-defined phrases we coded positive,
the false positive rates, i.e., 
portion of all truly-undefined phrases we coded positive,
and the number of entries discovered.
Upon performing these experiments,
the average of the ten trials
is taken for each of the three parameters,
for a number of short list lengths
(scanning $1,000$ $\log$-spaced lengths),
and plotted as a 
receiver operating characteristic
(ROC) curve 
(see Figs.~\ref{fig:twitter.crossval}--\ref{fig:gutenberg.crossval}).
We also note that each is also presented with
its area under curve (AUC),
which measures the accuracy
of the expanding-list classifier 
as a whole.

\begin{table}[t!]
  \begin{center}
    \begin{tabular}{|c|c|c|c|c|c|}
      \hline
      & \textbf{Corpus} & \textbf{2-gram} & \textbf{3-gram} & \textbf{4-gram} & \textbf{5-gram} \\\hline
      \multirow{5}{*}{\rotatebox[origin=c]{90}{\bf Cross-val}}
      & Twitter & 4.22 (0.40) & 1.11 (0.30) & 0.90 (0.10) & 1.49 (0) \\
      & NYT & 4.97 (0.30)& 0.36 (0.50) & 0.59 (0.10) & 1.60 (0) \\
      & Lyrics & 3.52 (0.50) & 1.76 (0.40) & 0.78 (0) & 0.48 (0) \\
      & Wikipedia & 5.06 (0.20) & 0.46 (0.80) & 1.94 (0.20) & 1.54 (0) \\
      & eBooks & 3.64 (0.30) & 1.86 (0.30) & 0.59 (0.60) & 0.90 (0.10) \\
      \hline\hline
      & \textbf{Corpus} & \textbf{2-gram} & \textbf{3-gram} & \textbf{4-gram} & \textbf{5-gram} \\\hline
      \multirow{5}{*}{\rotatebox[origin=c]{90}{\bf Live exp.}}
      & Twitter & 6(0) & 4 (0) & 5 (0) & 5 (0) \\
      & NYT & 5 (0) & 0 (0) & 2 (0) & 1 (0) \\
      & Lyrics & 3 (0) & 1 (0) & 3 (0) & 1 (0) \\
      & Wikipedia & 0 (0) & 1 (0) & 1 (0) & 2 (0) \\
      & eBooks & 2 (0) & 1 (0) & 3 (0) & 6 (1) \\
      \hline
    \end{tabular}
    \caption{
      Summarizing our results
      from the cross-validation procedure \textbf{(Above)},
      we present the mean numbers of missing 
      entries discovered
      when $20$ guesses were made
      for $N$-grams/phrases of lengths $2$, $3$, $4$, and $5$, each.
      For each of the $5$ large corpora (see Materials and Methods)
      we make predictions according our likelihood filter,
      and according to frequency (in parentheses)
      as a baseline.
      When considering the $2$-grams
      (for which the most definition information exists),
      short lists of $20$
      rendered up to $25\%$ correct predictions 
      on average by the definition likelihood,
      as opposed to the frequency ranking,
      by which no more than $2.5\%$ could be expected.
      We also summarize the results to-date
      from the live experiment \textbf{(Below)}
      (updated February 19, 2015),
      and present the numbers of missing 
      entries correctly discovered
      on the Wiktionary
      (i.e., reference added since July 1, 2014,
      when the dictionary's data was accessed)
      by the $20$-phrase shortlists 
      produced in our experiments for both
      the likelihood and frequency (in parentheses) filters.
      Here we see that all of the corpora
      analyzed were generative of phrases,
      with Twitter far and away being the most productive,
      and the reference corpus Wikipedia the least so.
    }
    \label{tab:numsDisc}
  \end{center}
\end{table}

\section{Results and discussion}

Before observing output from our model
we take the time to perform a cross-validation
($10$-fold), 
and compare our context filter to 
a sort by frequency alone.
From this we have found
that our likelihood filter 
renders missing entries much more efficiently than by frequency
(see Tab.~\ref{tab:numsDisc}, and Figs.~\ref{fig:twitter.crossval}--\ref{fig:gutenberg.crossval}),
already discovering missing entries
from short lists of as little as twenty 
(see the insets of Figs.~\ref{fig:twitter.crossval}--\ref{fig:gutenberg.crossval}
as well as Tabs.~\ref{tab:numsDisc},~\ref{tab:twitter},~and~\ref{tab:times}--\ref{tab:gutenberg}).
As such we adhere to this standard,
and only publish short lists of 
$20$ predictions per corpus
per phrase lengths $2$--$5$.
In parallel, we also present 
phrase frequency-generated short-lists
for comparison.

\begin{table*}
  \begin{center}
    \scalebox{0.93}{
      \begin{tabular}{|c|c|c|c|c|c|}
        \hline
        & \textbf{rank} & \textbf{2-gram} & \textbf{3-gram} & \textbf{4-gram} & \textbf{5-gram} \\\hline
        \multirow{20}{*}{\rotatebox[origin=c]{90}{\bf definition likelihood}}
        & 1 & buenos noches & knock it out & in the same time & actions speak louder then words \\
        & 2 & \color{red}{north york} & \color{red}{walk of fame} & on the same boat & \color{red}{no sleep for the wicked} \\
        & 3 & last few & piece of mind & about the same time & every once and a while \\
        & 4 & \color{red}{holy hell} & seo-search engine optimization & around the same time & to the middle of nowhere \\
        & 5 & good am & puta q pariu & at da same time & \color{red}{come to think about it} \\
        & 6 & \color{red}{going away} & \color{red}{who the heck} & wat are you doing & dont let the bedbugs bite \\
        & 7 & right up & take it out & wtf are you doing & you get what i mean \\
        & 8 & go sox & fim de mundo & why are you doing & you see what i mean \\
        & 9 & going well & note to all & hell are you doing & you know who i mean \\
        & 10 & due out & in the moment & better late then never & no rest for the weary \\
        & 11 & last bit & note to myself & here i go again & as long as i know \\
        & 12 & \color{red}{go far} & check it here & \color{red}{every now and again} & as soon as i know \\
        & 13 & right out & check it at & what were you doing & \color{red}{going out on a limb} \\
        & 14 & fuck am & check it http & \color{red}{was it just me} & \color{red}{give a person a fish} \\
        & 15 & holy god & check it now & here we are again & at a lost for words \\
        & 16 & rainy morning & check it outhttp & \color{red}{keeping an eye out} & de una vez por todas \\
        & 17 & \color{red}{picked out} & \color{red}{why the heck} & what in the butt & onew kids on the block \\
        & 18 & south coast & memo to self & de vez em qdo & twice in a blue moon \\
        & 19 & every few & reminder to self & \color{red}{giving it a try} & just what the dr ordered \\
        & 20 & \color{red}{picking out} & \color{red}{how the heck} & \color{red}{pain in my ass} & \color{red}{as far as we know} \\
        \hline\hline
        & \textbf{rank} & \textbf{2-gram} & \textbf{3-gram} & \textbf{4-gram} & \textbf{5-gram} \\\hline
        \multirow{20}{*}{\rotatebox[origin=c]{90}{\bf frequency}}
        & 1 & in the & new blog post & i just took the & i favorited a youtube video \\
        & 2 & i just & i just took & e meu resultado foi & i uploaded a youtube video \\
        & 3 & of the & live on http & other people at http & just joined a video chat \\
        & 4 & on the & i want to & check this video out & fiddling with my blog post \\
        & 5 & i love & i need to & just joined a video & joined a video chat with \\
        & 6 & i have & i have a & a day using http & i rated a youtube video \\
        & 7 & i think & quiz and got & on my way to & i just voted for http \\
        & 8 & to be & thanks for the & favorited a youtube video & this site just gave me \\
        & 9 & i was & what about you & i favorited a youtube & add a \#twibbon to your \\
        & 10 & if you & i think i & free online adult dating & the best way to get \\
        & 11 & at the & i have to & a video chat with & just changed my twitter background \\
        & 12 & have a & looking forward to & uploaded a youtube video & a video chat at http \\
        & 13 & to get & acabo de completar & i uploaded a youtube & photos on facebook in the \\
        & 14 & this is & i love it & video chat at http & check it out at http \\
        & 15 & and i & a youtube video & what do you think & own video chat at http \\
        & 16 & but i & to go to & i am going to & s channel on youtube http \\
        & 17 & are you & of the day & if you want to & and won in \#mobsterworld http \\
        & 18 & it is & what'll you get & i wish i could & live stickam stream at http \\
        & 19 & i need & my daily twittascope & just got back from & on facebook in the album \\
        & 20 & it was & if you want & thanks for the rt & added myself to the http \\
        \hline
      \end{tabular}
    }
    \caption{
      With data taken from the Twitter corpus,
      we present the top $20$ unreferenced phrases
      considered for definition (in the live experiment)
      from each of the $2$, $3$, $4$, and $5$-gram 
      likelihood filters \textbf{(Above)},
      and frequency filters \textbf{(Below)}.
      From this corpus we note the juxtaposition
      of highly idiomatic expressions by the likelihood filter
      (like ``holy hell''),      
      with the domination of the frequency filters
      by semi-automated content.
      The phrase ``holy hell'' is an
      example of the model's
      success with this corpus,
      as it achieved definition (February $8^\text{th}$, 2015)
      concurrently with the preparation of this manuscript
      (several months after the Wiktionary's data was accessed in July, 2014).
    }
    \label{tab:twitter}
  \end{center}
\end{table*}

In addition to listing them in the appendices,
we have presented the results
of our experiment from across the $5$ 
large, disparate corpora
on the Wiktionary in a pilot program, 
where we are tracking the success of the filters~\protect{\footnote{Track the potential missing entries 
that we have proposed: 
\url{https://en.wiktionary.org/wiki/User:Jakerylandwilliams/Potential_missing_entries}}}.
Looking at the lexical tables,
where defined phrases are highlighted in red,
we can see that many of the predictions
by the likelihood filter
(especially those obtained from the Twitter corpus)
have already been defined in the Wiktionary
following our recommendation
(as of Feb. 19th 2015)
since we accessed its data
in July of 2014~\cite{wiktionary2014}.
We also summarize these results from the live
experiment in Tab.~\ref{tab:numsDisc}.

Looking at the lexical tables more closely, 
we note that all corpora present 
highly idiomatic expressions
under the likelihood filter,
many of which are variants of existing
idiomatic phrases that will 
likely be granted inclusion
into the dictionary through redirects
or alternative-forms listings.
To name a few, the 
Twitter (Tab.~\ref{tab:twitter}), 
Times (Tab.~\ref{tab:times}), 
and Lyrics (Tab.~\ref{tab:lyrics})
corpora consistently predict
large families derived from phrases like
``at the same time'', and ``you know what i mean'',
while the eBooks and Wikipedia corpora
predict families derived from phrases like
``on the other hand'', and ``at the same time''.
In general we see no such structure
or predictive power emerge
from the frequency filter.

We also observe that from those corpora which are
less pure of English context
(namely, the eBooks, Lyrics, and Twitter corpora),
extra-English expressions have crept in.
This highlights an important feature of the 
likelihood filter---it does not intrinsically rely on
the syntax or grammar of the language to which it is applied,
beyond the extent to which syntax and
grammar effect the shapes of collocations.
For example, the eBooks predict
(see Tab.~\ref{tab:gutenberg})
the undefined French phrase
``tu ne sais pas'', or ``you do not know'',
which is a syntactic variant of the English-Wiktionary defined
French, ``je ne sais pas'', meaning ``i do not know''.
Seeing this, we note that it would be straightforward 
to construct a likelihood filter with a language indicator norm 
to create an alternative framework for language identification.

There are also a fair number of
phrases predicted by the likelihood filter
which in fact are spelling errors, typos, 
and grammatical errors.
In terms of the context model,
these erroneous forms are quite near to those
defined in the dictionary, 
and so rise in the short lists generated from the
less-well edited corpora, e.g., 
``actions~speak~louder~\textit{then}~words''
in the Twitter corpus.
This then seems to indicate the potential
for the likelihood filter to be
integrated into auto-correct algorithms,
and further points to the possibility
of constructing syntactic indicator
norms of phrases, making estimations of
tenses and parts of speech
(whose data is also available from the Wiktionary~\cite{wiktionary2014})
possible through application of the model
in precisely the same manner presented in Sec.~\ref{sec:deflik}.
Regardless of the future applications,
we have developed and presented a 
novel, powerful, and scalable
MWE extraction technique.

\clearpage

\onecolumngrid
\appendix

\section{Cross-validation results for missing entry detection}
\subsection{The New York Times}
\begin{figure*}[b!]
  \includegraphics[width=0.95\textwidth]{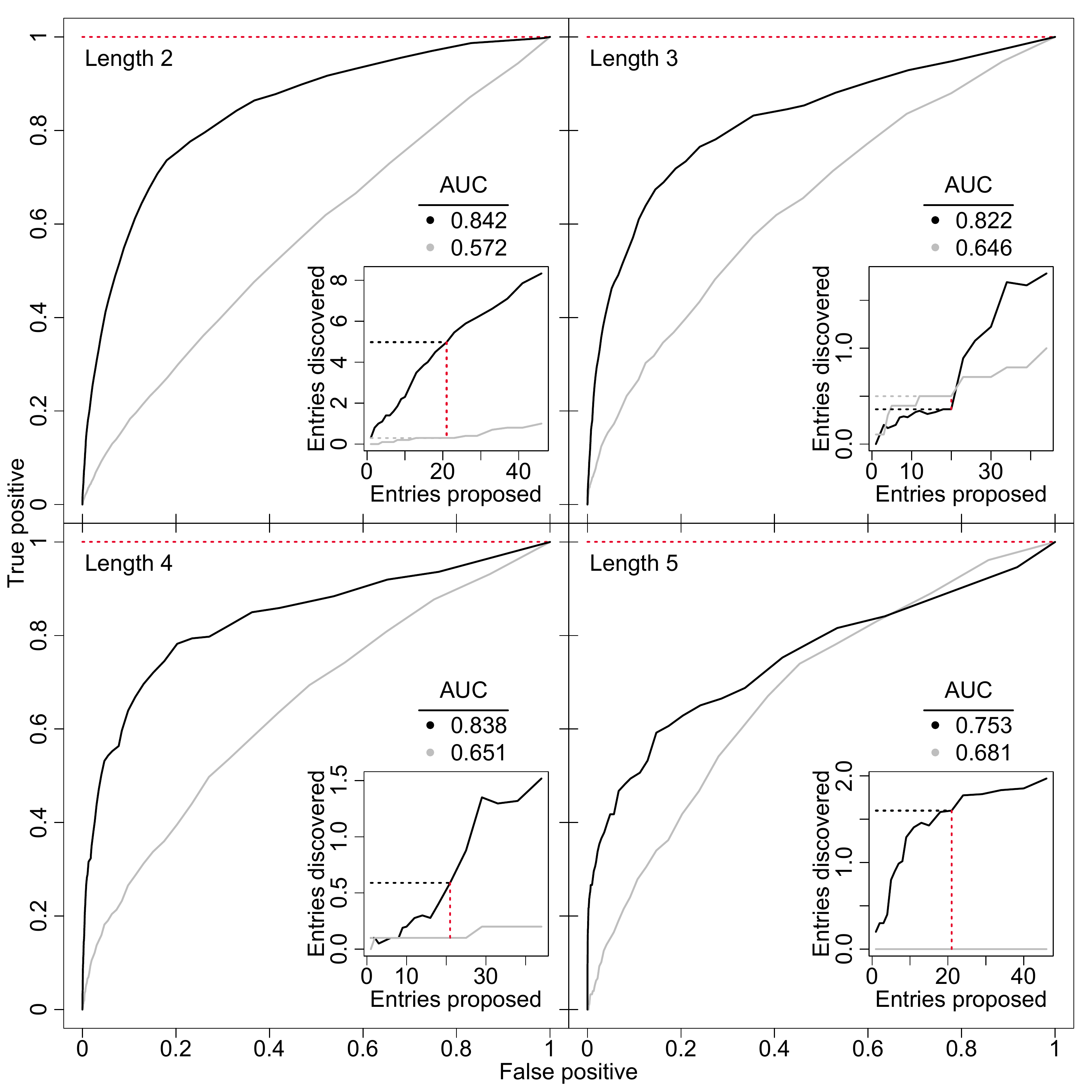}
  \caption{
    With data taken from the NYT corpus, 
    we present ($10$-fold) cross-validation results
    for the filtration procedures.
    For each of the lengths $2$, $3$, $4$, and $5$,
    we show the ROC curves (\textbf{Main Axes}),
    comparing true and false positive rates
    for both the likelihood filters (black),
    and for the frequency filters (gray).
    There, we see increased performance 
    in the likelihood classifiers
    (except possibly for length $5$),
    which is reflected in the AUCs
    (where an AUC of $1$ indicates a perfect classifier).
    We also monitor the average number of 
    missing entries discovered
    as a function of the number of entries proposed (\textbf{Insets}),
    for each length.
    There, the horizontal dotted lines indicate 
    the average numbers
    of missing entries discovered for both the
    likelihood filters (black)
    and for the frequency filters (gray)
    when short lists of $20$ phrases were taken 
    (red dotted vertical lines).
    From this we see an indication
    that even the $5$-gram likelihood filter is
    effective at detecting missing entries in short lists,
    while the frequency filter is not.
  }
  \label{fig:times.crossval}
\end{figure*}

\newpage
\subsection{Music Lyrics}
\begin{figure*}[b!]
  \includegraphics[width=\textwidth]{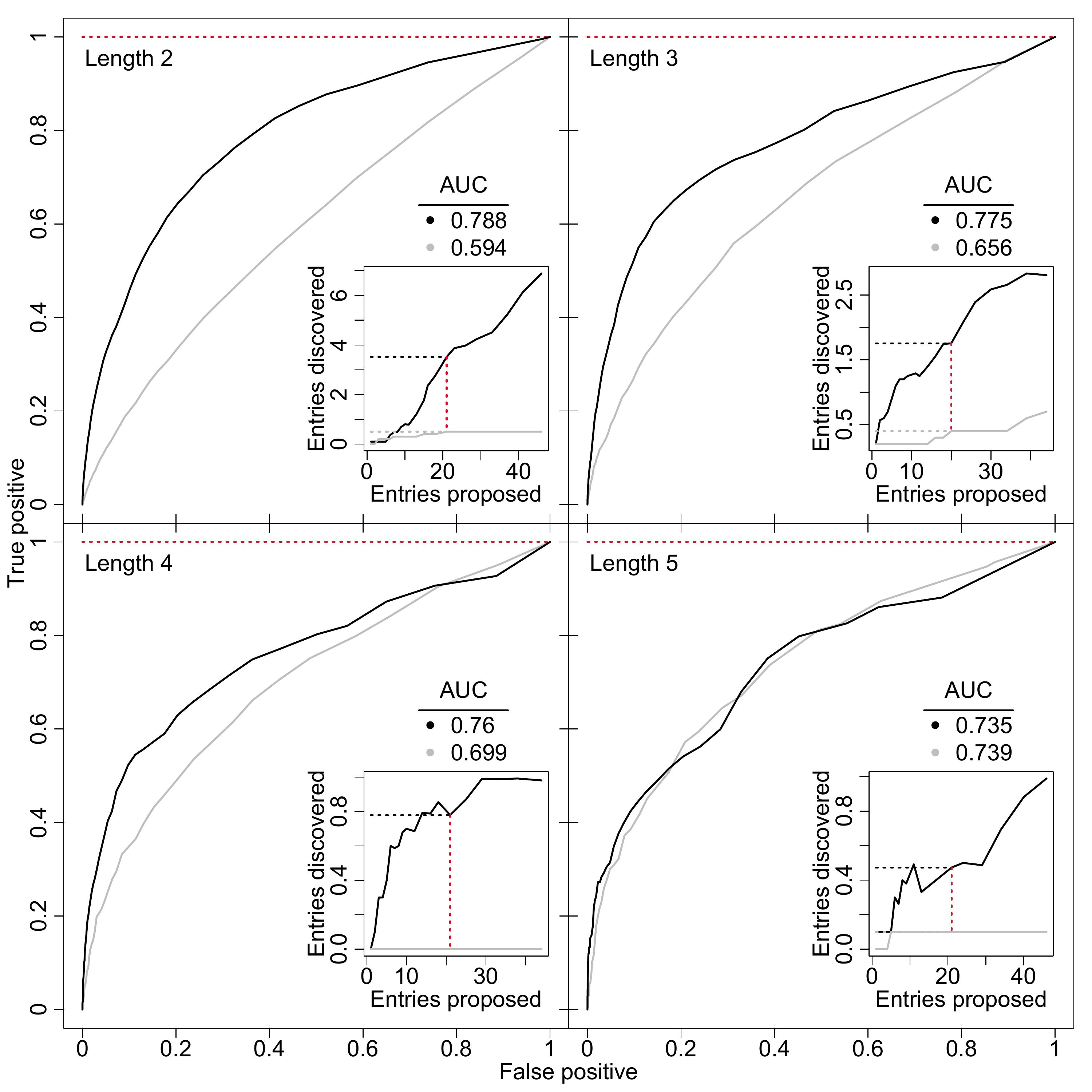}
  \caption{
    With data taken from the Lyrics corpus,
    we present ($10$-fold) cross-validation results
    for the filtration procedures.
    For each of the lengths $2$, $3$, $4$, and $5$,
    we show the ROC curves (\textbf{Main Axes}),
    comparing true and false positive rates
    for both the likelihood filters (black),
    and for the frequency filters (gray).
    There, we see increased performance 
    in the likelihood classifiers,
    which is reflected in the AUCs
    (where an AUC of $1$ indicates a perfect classifier).
    We also monitor the average number of 
    missing entries discovered
    as a function of the number of entries proposed (\textbf{Insets}),
    for each length.
    There, the horizontal dotted lines indicate 
    the average numbers
    of missing entries discovered for both the
    likelihood filters (black)
    and for the frequency filters (gray),
    when short lists of $20$ phrases were taken 
    (red dotted vertical lines).
    Here we can see that 
    it may have been advantageous
    to construct a slightly longer $3$ and $4$-gram lists.
  }
  \label{fig:lyrics.crossval}
\end{figure*}

\newpage
\subsection{English Wikipedia}
\begin{figure*}[b!]
  \includegraphics[width=\textwidth]{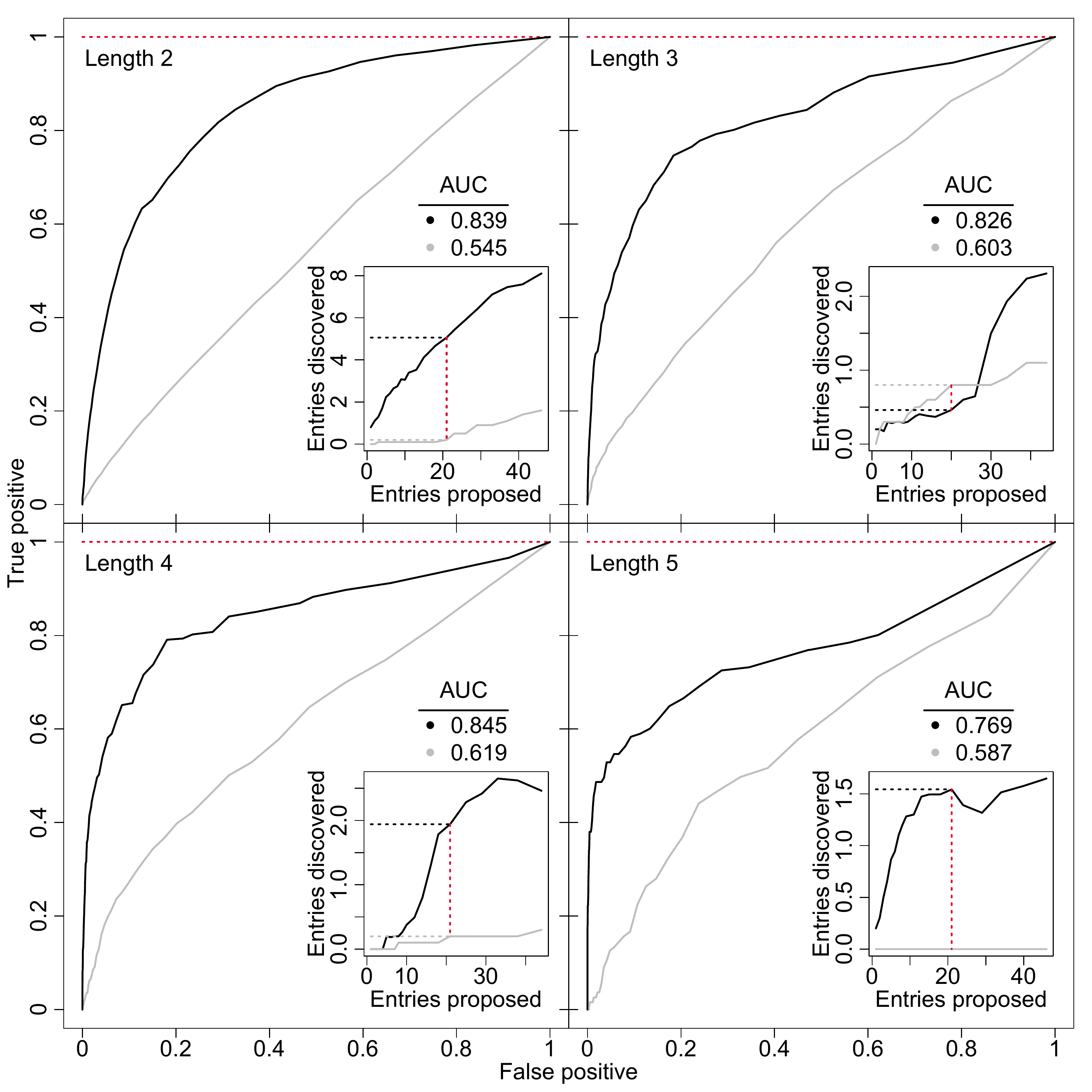}
  \caption{
    With data taken from the Wikipedia corpus,
    we present ($10$-fold) cross-validation results
    for the filtration procedures.
    For each of the lengths $2$, $3$, $4$, and $5$,
    we show the ROC curves (\textbf{Main Axes}),
    comparing true and false positive rates
    for both the likelihood filters (black),
    and for the frequency filters (gray).
    There, we see increased performance 
    in the likelihood classifiers,
    which is reflected in the AUCs
    (where an AUC of $1$ indicates a perfect classifier).
    We also monitor the average number of 
    missing entries discovered
    as a function of the number of entries proposed (\textbf{Insets}),
    for each length.
    There, the horizontal dotted lines indicate 
    the average numbers
    of missing entries discovered for both the
    likelihood filters (black)
    and for the frequency filters (gray)
    when short lists of $20$ phrases were taken 
    (red dotted vertical lines).
    Here we can see that 
    it may have been advantageous
    to construct a slightly longer $3$ and $4$-gram lists.
  }
  \label{fig:wikipedia.crossval}
\end{figure*}

\newpage
\subsection{Project Gutenberg eBooks}
\begin{figure*}[b!]
  \includegraphics[width=\textwidth]{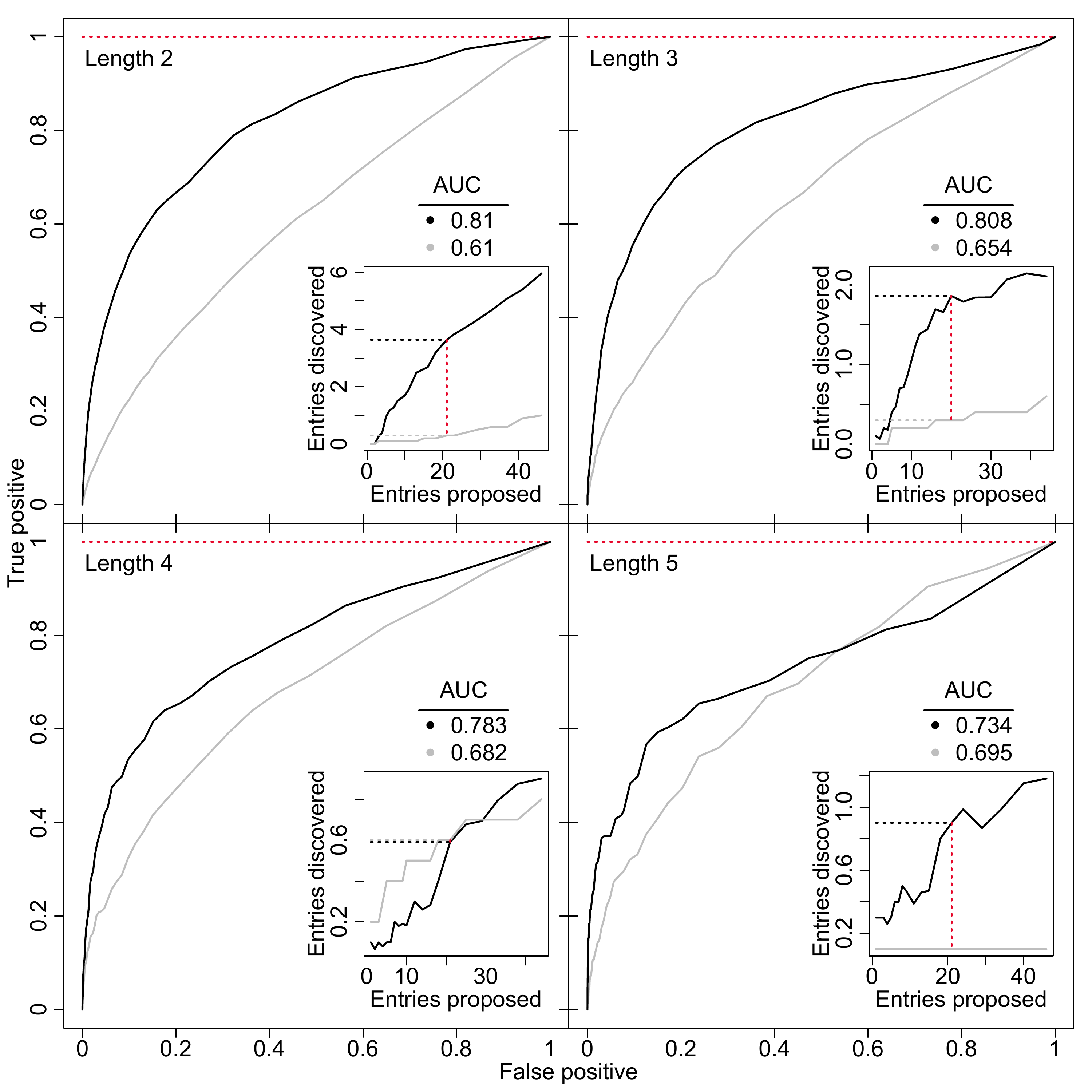}
  \caption{
    With data taken from the eBooks corpus,
    we present ($10$-fold) cross-validation results
    for the filtration procedures.
    For each of the lengths $2$, $3$, $4$, and $5$,
    we show the ROC curves (\textbf{Main Axes}),
    comparing true and false positive rates
    for both the likelihood filters (black),
    and for the frequency filters (gray).
    There, we see increased performance 
    in the likelihood classifiers,
    which is reflected in the AUCs
    (where an AUC of $1$ indicates a perfect classifier).
    We also monitor the average number of 
    missing entries discovered
    as a function of the number of entries proposed (\textbf{Insets}),
    for each length.
    There, the horizontal dotted lines indicate 
    the average numbers
    of missing entries discovered for both the
    likelihood filters (black)
    and for the frequency filters (gray)
    when short lists of $20$ phrases were taken 
    (red dotted vertical lines).
    Here we can see that the 
    power of the $4$-gram model
    does not show itself until 
    longer lists are considered.
  }
  \label{fig:gutenberg.crossval}
\end{figure*}

\newpage
\section {Tables of potential missing entries}
\subsection{The New York Times}
\begin{table*}[b!]
  \begin{center}
    \scalebox{0.935}{
      \begin{tabular}{|c|c|c|c|c|c|}
        \hline
        & \textbf{rank} & \textbf{2-gram} & \textbf{3-gram} & \textbf{4-gram} & \textbf{5-gram} \\\hline
        \multirow{20}{*}{\rotatebox[origin=c]{90}{\bf definition likelihood}}
        & 1 & prime example & as united states & in the same time & when push came to shove \\
        & 2 & going well & in united states & about the same time & nat. ocean. and atm. admin. \\
        & 3 & south jersey & by united states & around the same time & all's well that ends well' \\
        & 4 & north jersey & eastern united states & during the same time & you see what i mean \\
        & 5 & united front & first united states & roughly the same time & so far as i know \\
        & 6 & go well & a united states & return to a boil & take it or leave it' \\
        & 7 & gulf states & to united states & \color{red}{every now and again} & gone so far as to \\
        & 8 & united germany & for united states & at the very time & love it or leave it \\
        & 9 & dining out & senior united states & nowhere to be seen & as far as we're concerned \\
        & 10 & north brunswick & of united states & for the long run & as bad as it gets \\
        & 11 & \color{red}{go far} & from united states & over the long run & as far as he's concerned \\
        & 12 & \color{red}{going away} & is a result & why are you doing & days of wine and roses' \\
        & 13 & there all & and united states & in the last minute & \color{red}{as far as we know} \\
        & 14 & \color{red}{picked out} & with united states & to the last minute & state of the county address \\
        & 15 & go all & that united states & until the last minute & state of the state address \\
        & 16 & this same & two united states & remains to be done & state of the city address \\
        & 17 & \color{red}{civil court} & its united states & \color{red}{turn of the screw} & just a matter of time \\
        & 18 & good example & assistant united states & turn of the last & be a matter of time \\
        & 19 & \color{red}{this instance} & but united states & turn of the millennium & for the grace of god \\
        & 20 & how am & western united states & once upon a mattress & short end of the market \\
        \hline\hline
        & \textbf{rank} & \textbf{2-gram} & \textbf{3-gram} & \textbf{4-gram} & \textbf{5-gram} \\\hline
        \multirow{20}{*}{\rotatebox[origin=c]{90}{\bf frequency}}
        & 1 & of the & one of the & in the united states & at the end of the \\
        & 2 & in the & in new york & for the first time & because of an editing error \\
        & 3 & he said & the new york & the new york times & the new york stock exchange \\
        & 4 & and the & some of the & in new york city & for the first time in \\
        & 5 & for the & part of the & at the end of & he is survived by his \\
        & 6 & at the & of new york & the end of the & is survived by his wife \\
        & 7 & in a & president of the & a spokesman for the & an initial public offering of \\
        & 8 & to be & the end of & at the university of & by the end of the \\
        & 9 & with the & there is a & one of the most & the end of the year \\
        & 10 & that the & director of the & of the united states & the securities and exchange commission \\
        & 11 & it is & it was a & a member of the & for the first time since \\
        & 12 & from the & according to the & the rest of the & for students and the elderly \\
        & 13 & she said & in the last & at the age of & beloved wife of the late \\
        & 14 & by the & the white house & to the united states & he said in an interview \\
        & 15 & it was & in the united & in lieu of flowers & the dow jones industrial average \\
        & 16 & as a & the university of & executive director of the & the executive director of the \\
        & 17 & he was & there is no & the united states and & tonight and tomorrow night at \\
        & 18 & is a & it is a & is one of the & in the last two years \\
        & 19 & with a & the first time & of the new york & in the new york times \\
        & 20 & and a & in the first & by the end of & in the last few years \\
        \hline
      \end{tabular}
    }
    \caption{
      With data taken from the NYT corpus,
      we present the top $20$ unreferenced phrases
      considered for definition (in the live experiment)
      from each of the $2$, $3$, $4$, and $5$-gram 
      likelihood filters \textbf{(Above)},
      and frequency filters \textbf{(Below)}.
      From this corpus we note the juxtaposition
      of highly idiomatic expressions by the likelihood filter
      (like ``united front''),      
      with the domination of the frequency filters
      by structural elements of rigid content
      (e.g., the obituaries).
      The phrase ``united front''
      is an example 
      of the model's success with this corpus,
      as it's coverage in a Wikipedia article
      began in 2006,
      describing the general Marxist tactic extensively.
      We also note that we have
      abbreviated 
      ``national oceanographic and atmospheric administration''
      \textbf{(Above)}, for brevity.      
    }
    \label{tab:times}
  \end{center}
\end{table*}

\newpage
\subsection{Music Lyrics}
\begin{table*}[b!]
  \begin{center}
    \scalebox{0.95}{
      \begin{tabular}{|c|c|c|c|c|c|}
        \hline
        & \textbf{rank} & \textbf{2-gram} & \textbf{3-gram} & \textbf{4-gram} & \textbf{5-gram} \\\hline
        \multirow{20}{*}{\rotatebox[origin=c]{90}{\bf definition likelihood}}
        & 1 & uh ha & now or later & one of a million & when push come to shove \\
        & 2 & come aboard & change of mind & made up your mind & come hell of high water \\
        & 3 & \color{red}{strung up} & \color{red}{over and done} & every now and again & you see what i mean \\
        & 4 & fuck am & forth and forth & \color{red}{make up my mind} & you know that i mean \\
        & 5 & \color{red}{iced up} & in and down & son of the gun & until death do us part \\
        & 6 & merry little & now and ever & cry me a river-er & that's a matter of fact \\
        & 7 & get much & off the air & have a good day & it's a matter of fact \\
        & 8 & da same & on and go & on way or another & what goes around comes back \\
        & 9 & \color{red}{messed around} & check it check & for the long run & you reap what you sew \\
        & 10 & old same & stay the fuck & feet on solid ground & to the middle of nowhere \\
        & 11 & used it & set the mood & feet on the floor & actions speak louder than lies \\
        & 12 & uh yeah & night to day & between you and i & u know what i mean \\
        & 13 & uh on & day and every & \color{red}{what in the hell} & ya know what i mean \\
        & 14 & fall around & meant to stay & why are you doing & you'll know what i mean \\
        & 15 & come one & in love you & you don't think so & you'd know what i mean \\
        & 16 & out much & upon the shelf & for better or for & y'all know what i mean \\
        & 17 & last few & up and over & once upon a dream & baby know what i mean \\
        & 18 & used for & check this shit & over and forever again & like it or leave it \\
        & 19 & number on & to the brink & knock-knock-knockin' on heaven's door & i know what i mean \\
        & 20 & come prepared & on the dark & once upon a lifetime & \color{red}{ain't no place like home} \\
        \hline\hline
        & \textbf{rank} & \textbf{2-gram} & \textbf{3-gram} & \textbf{4-gram} & \textbf{5-gram} \\\hline
        \multirow{20}{*}{\rotatebox[origin=c]{90}{\bf frequency}}
        & 1 & in the & i want to & la la la la & la la la la la \\
        & 2 & and i & la la la & i don't want to & na na na na na \\
        & 3 & i don't & i want you & na na na na & on and on and on \\
        & 4 & on the & you and me & in love with you & i want you to know \\
        & 5 & if you & i don't want & i want you to & don't know what to do \\
        & 6 & to me & i know you & i don't know what & oh oh oh oh oh \\
        & 7 & to be & i need you & i don't know why & da da da da da \\
        & 8 & i can & and i know & oh oh oh oh & do do do do do \\
        & 9 & and the & i don't wanna & i want to be & one more chance at love \\
        & 10 & but i & i got a & know what to do & i don't want to be \\
        & 11 & of the & i know that & what can i do & in the middle of the \\
        & 12 & i can't & you know i & yeah yeah yeah yeah & i don't give a fuck \\
        & 13 & for you & i can see & you don't have to & yeah yeah yeah yeah yeah \\
        & 14 & when i & and i don't & i close my eyes & i don't know what to \\
        & 15 & you can & in your eyes & you want me to & all i want is you \\
        & 16 & i got & and if you & you make me feel & you know i love you \\
        & 17 & in my & the way you & i just want to & the middle of the night \\
        & 18 & all the & na na na & da da da da & the rest of my life \\
        & 19 & i want & don't you know & if you want to & no no no no no \\
        & 20 & that i & this is the & come back to me & at the end of the \\
        \hline
      \end{tabular}
    }
    \caption{
      With data taken from the Lyrics corpus,
      we present the top $20$ unreferenced phrases
      considered for definition (in the live experiment)
      from each of the $2$, $3$, $4$, and $5$-gram 
      likelihood filters \textbf{(Above)},
      and frequency filters \textbf{(Below)}.
      From this corpus we note the juxtaposition
      of highly idiomatic expressions by the likelihood filter
      (like ``iced up''),      
      with the domination of the frequency filters
      by various onomatopoeiae.
      The phrase ``iced up''
      is an example 
      of the model's
      success with this corpus,
      having had definition in the Urban Dictionary
      since 2003, indicating that one is 
      ``covered in diamonds''.
      Further, though this phrase
      does have a variant that is defined in the Wiktionary (as early as 2011)---``iced out''---we
      note that the reference is also made in the Urban Dictionary
      (as early as 2004),
      where the phrase has distinguished meaning
      for one that is so bedecked---ostentatiously.
    }
    \label{tab:lyrics}
  \end{center}
\end{table*}

\newpage
\subsection{English Wikipedia}
\begin{table*}[b!]
  \begin{center}
    \scalebox{0.975}{
      \begin{tabular}{|c|c|c|c|c|c|}
        \hline
        & \textbf{rank} & \textbf{2-gram} & \textbf{3-gram} & \textbf{4-gram} & \textbf{5-gram} \\\hline\hline
        \multirow{20}{*}{\rotatebox[origin=c]{90}{\bf definition likelihood}}
        & 1 & new addition & in respect to & in the other hand & the republic of the congo \\
        & 2 & african states & as united states & people's republic of poland & so far as i know \\
        & 3 & less well & was a result & people's republic of korea & going as far as to \\
        & 4 & south end & \color{red}{walk of fame} & in the same time & gone so far as to \\
        & 5 & dominican order & central united states & the republic of congo & went as far as to \\
        & 6 & united front & in united states & at this same time & goes as far as to \\
        & 7 & same-sex couples & eastern united states & at that same time & the federal republic of yugoslavia \\
        & 8 & baltic states & first united states & approximately the same time & state of the nation address \\
        & 9 & to york & a united states & about the same time & \color{red}{as far as we know} \\
        & 10 & new kingdom & under united states & around the same time & just a matter of time \\
        & 11 & east carolina & to united states & during the same time & due to the belief that \\
        & 12 & due east & of united states & roughly the same time & as far as i'm aware \\
        & 13 & united church & southern united states & ho chi minh trail & due to the fact it \\
        & 14 & quarter mile & southeastern united states & lesser general public license & due to the fact he \\
        & 15 & end date & southwestern united states & in the last minute & due to the fact the \\
        & 16 & so well & and united states & on the right hand & \color{red}{as a matter of course} \\
        & 17 & olympic medalist & th united states & on the left hand & as a matter of policy \\
        & 18 & at york & western united states & once upon a mattress & as a matter of principle \\
        & 19 & go go & for united states & o caetano do sul & or something to that effect \\
        & 20 & teutonic order & former united states & \color{red}{turn of the screw} & as fate would have it \\
        \hline\hline
        & \textbf{rank} & \textbf{2-gram} & \textbf{3-gram} & \textbf{4-gram} & \textbf{5-gram} \\\hline\hline
        \multirow{20}{*}{\rotatebox[origin=c]{90}{\bf frequency}}
        & 1 & of the & one of the & in the united states & years of age or older \\
        & 2 & in the & part of the & at the age of & the average household size was \\
        & 3 & and the & the age of & a member of the & were married couples living together \\
        & 4 & on the & the end of & under the age of & from two or more races \\
        & 5 & at the & according to the & the end of the & at the end of the \\
        & 6 & for the & may refer to & at the end of & the median income for a \\
        & 7 & he was & member of the & as well as the & the result of the debate \\
        & 8 & it is & the university of & years of age or & of it is land and \\
        & 9 & with the & in the early & of age or older & the racial makeup of the \\
        & 10 & as a & a member of & the population density was & has a total area of \\
        & 11 & it was & in the united & the median age was & the per capita income for \\
        & 12 & from the & he was a & as of the census & and the average family size \\
        & 13 & the first & of the population & households out of which & and the median income for \\
        & 14 & as the & was born in & one of the most & the average family size was \\
        & 15 & was a & end of the & people per square mile & had a median income of \\
        & 16 & in a & in the late & at the university of & of all households were made \\
        & 17 & to be & in addition to & was one of the & at an average density of \\
        & 18 & one of & it is a & for the first time & males had a median income \\
        & 19 & during the & such as the & the result of the & housing units at an average \\
        & 20 & with a & the result was & has a population of & made up of individuals and \\
        \hline\hline
      \end{tabular}
    }
    \caption{
      With data taken from the Wikipedia corpus,
      we present the top $20$ unreferenced phrases
      considered for definition (in the live experiment)
      from each of the $2$, $3$, $4$, and $5$-gram 
      likelihood filters \textbf{(Above)},
      and frequency filters \textbf{(Below)}.
      From this corpus we note the juxtaposition
      of highly idiomatic expressions by the likelihood filter
      (like ``same-sex couples''),      
      with the domination of the frequency filters
      by highly-descriptive structural
      text from the presentations of
      demographic and numeric data.
      The phrase
      ``same-sex couples''
      is an example 
      of the model's
      success with this corpus,
      and appears largely because of the existence distinct phrases
      ``same-sex marriage'' and 
      ``married couples''
      with definition in the Wiktionary.
    }
    \label{tab:wikipedia}
  \end{center}
\end{table*}

\newpage
\subsection{Project Gutenberg eBooks}
\begin{table*}[b!]
  \begin{center}
    \scalebox{0.975}{
      \begin{tabular}{|c|c|c|c|c|c|}
        \hline
        & \textbf{rank} & \textbf{2-gram} & \textbf{3-gram} & \textbf{4-gram} & \textbf{5-gram} \\\hline\hline
        \multirow{20}{*}{\rotatebox[origin=c]{90}{\bf definition likelihood}}
        & 1 & go if & \color{red}{by and bye} & i ask your pardon & \color{red}{handsome is that handsome does} \\
        & 2 & come if & purchasing power equivalent & i crave your pardon & for the grace of god \\
        & 3 & able man & of the contrary & with the other hand & \color{red}{be that as it might} \\
        & 4 & at york & quite the contrary & \color{red}{upon the other hand} & \color{red}{be that as it will} \\
        & 5 & going well & of united states & about the same time & up hill and down hill \\
        & 6 & there once & so well as & and the same time & \color{red}{come to think about it} \\
        & 7 & go well & at a rate & \color{red}{every now and again} & is no place like home \\
        & 8 & so am & point of fact & tu ne sais pas & for the love of me \\
        & 9 & go all & as you please & quarter of an inch & so far as i'm concerned \\
        & 10 & \color{red}{picked out} & so soon as & quarter of an ounce & you know whom i mean \\
        & 11 & very same & it a rule & quarter of an hour's & you know who i mean \\
        & 12 & come all & so to bed & qu'il ne fallait pas & upon the face of it \\
        & 13 & look well & of a hurry & to the expense of & you understand what i mean \\
        & 14 & there all & at the rate & be the last time & you see what i mean \\
        & 15 & how am & such a hurry & and the last time & by the grace of heaven \\
        & 16 & \color{red}{going away} & just the way & was the last time & by the grace of the \\
        & 17 & going forth & it all means & is the last time & don't know what i mean \\
        & 18 & get much & you don't know & so help me heaven & \color{red}{be this as it may} \\
        & 19 & why am & greater or less & \color{red}{make up my mind} & \color{red}{in a way of speaking} \\
        & 20 & this same & have no means & at the heels of & or something to that effect \\
        \hline\hline
        & \textbf{rank} & \textbf{2-gram} & \textbf{3-gram} & \textbf{4-gram} & \textbf{5-gram} \\\hline\hline
        \multirow{20}{*}{\rotatebox[origin=c]{90}{\bf frequency}}
        & 1 & of the & one of the & for the first time & at the end of the \\
        & 2 & and the & it was a & at the end of & and at the same time \\
        & 3 & it was & there was a & of the united states & the other side of the \\
        & 4 & on the & out of the & the end of the & on the part of the \\
        & 5 & it is & it is a & the rest of the & distributed proofreading team at http \\
        & 6 & to be & i do not & one of the most & on the other side of \\
        & 7 & he was & it is not & on the other side & at the foot of the \\
        & 8 & at the & and it was & for a long time & percent of vote by party \\
        & 9 & for the & it would be & it seems to me & at the head of the \\
        & 10 & with the & he did not & it would have been & \color{red}{as a matter of course} \\
        & 11 & he had & there was no & as well as the & on the morning of the \\
        & 12 & by the & and in the & i am going to & for the first time in \\
        & 13 & he said & that he was & as soon as the & it seems to me that \\
        & 14 & in a & it was not & i should like to & president of the united states \\
        & 15 & with a & it was the & as a matter of & at the bottom of the \\
        & 16 & and i & that he had & on the part of & i should like to know \\
        & 17 & that the & there is no & the middle of the & but at the same time \\
        & 18 & of his & that it was & the head of the & at the time of the \\
        & 19 & i have & he had been & at the head of & had it not been for \\
        & 20 & and he & but it was & the edge of the & at the end of a \\
        \hline\hline
      \end{tabular}
    }
    \caption{
      With data taken from the eBooks corpus,
      we present the top $20$ unreferenced phrases
      considered for definition (in the live experiment)
      from each of the $2$, $3$, $4$, and $5$-gram 
      likelihood filters \textbf{(Above)},
      and frequency filters \textbf{(Below)}.
      From this corpus we note the juxtaposition
      of many highly idiomatic expresisons by the likelihood filter,
      with the domination of the frequency filters
      by highly-structural text.
      Here, since the texts are all 
      within the public domain,
      we see that this much less modern corpus
      is without the innovation
      present in the other,
      but that the likelihood filter
      does still extract many
      unreferenced variants
      of Wiktionary-defined idiomatic forms.
    }
    \label{tab:gutenberg}
  \end{center}
\end{table*}

\end{document}